# Modified Double-DQN: addressing stability

Shervin Halat
Computer Engineering Department
Amirkabir University of Technology
Tehran, Iran
shervin.halat@aut.ac.ir

Mohammad Mehdi Ebadzadeh
Computer Engineering Department
Amirkabir University of Technology
Tehran, Iran
ebadzadeh@aut.ac.ir

Kiana Amani
Industrial Engineering Department
Sharif University of Technology
Tehran, Iran
kiana.amani01@sharif.edu

***Abstract*—Inspired by Double Q-learning algorithm, the Double-DQN (DDQN) algorithm was originally proposed in order to address the overestimation issue in the original DQN algorithm. The DDQN has successfully shown both theoretically and empirically the importance of decoupling in terms of action evaluation and selection in computation of target values; although, all the benefits were acquired with only a simple adaption to DQN algorithm, minimal possible change as it was mentioned by the authors. Nevertheless, there seems a roll-back in the proposed algorithm of DDQN since the parameters of policy network are emerged again in the target value function which were initially withdrawn by DQN with the hope of tackling the serious issue of moving targets and the instability caused by it (i.e., by moving targets) in the process of learning. Therefore, in this paper three modifications to the DDQN algorithm are proposed with the hope of maintaining the performance in the terms of both stability and overestimation. These modifications are focused on the logic of decoupling the best action selection and evaluation in the target value function and the logic of tackling the moving targets issue. Each of these modifications have their own pros and cons compared to the others. The mentioned pros and cons mainly refer to the execution time required for the corresponding algorithm and the stability provided by the corresponding algorithm. Also, in terms of overestimation, none of the modifications seem to underperform compared to the original DDQN if not outperform it. With the intention of evaluating the efficacy of the proposed modifications, multiple empirical experiments along with theoretical experiments were conducted. The results obtained are represented and discussed in this article.***

## I. Introduction

By maximizing a cumulative future reward signal, reinforcement learning generally aims to reach optimal policies for sequential decision problems [1]. A popular approach to achieve such goal is through Q-learning, proposed by Watkins in 1989, which can be used to optimally solve Markov Decision Processes [2]. Performing poorly in stochastic MDPs owing to large overestimations of the action values introduced by maximization step over estimated action values, Q-learning has shown inclination toward overestimated values [1, 2]. In an early revision to Q-learning algorithm, overestimations had been imputed to insufficient flexibility of function approximation and noise [1]. Hence, a noble implementation through deep neural networks emerged as a modification to traditional Q-learning algorithm, namely Deep Q-learning Network or DQN [3, 4], with corresponding unrolled target value function as (1).

$$Y_t^Q \equiv R_{t+1} + \gamma Q\left(S_{t+1}, \underset{a}{\operatorname{argmax}}\, Q(S_{t+1}, a; \theta_t'); \theta_t'\right) \quad (1)$$

Moreover, in another revision to Q-learning algorithm, Double Q-learning, it has been argued that these overestimations result from a positive bias that is employed by Q-learning which opted to approximate and substitute maximum expected action value with maximum action value [2] whereby in [2] the well-known Double Q-learning algorithm introduced, through which the two processes of action selection and action evaluation within target value function became decoupled (2) thanks to the employment of two sets of value functions or, in other words, two sets of weights.

$$Y_t^{DoubleQ} \equiv R_{t+1} + \gamma Q\left(S_{t+1}, \underset{a}{\operatorname{argmax}}\, Q(S_{t+1}, a; \theta_t); \theta_t'\right) \quad (2)$$

Next, having fused the two aforementioned ideas (i.e., DQN and Double Q-learning) and incorporated them into Q-learning, a more stable algorithm, Double-DQN (DDQN) [1], was introduced mainly with the aim of reducing the overestimation.

The main idea of DDQN, inspired by Double Q-learning, is to reduce overestimations through detachment of action selection and action evaluation of the max operation in the target value function; however, in contrast to Double Q-learning, as [1] puts it, minimal possible changes to DQN as a means to incorporate Double Q-learning idea were considered, seemingly to keep DQN algorithm intact as much as possible in addition to not imposing additional computational overhead. Consequently, although not fully decoupled, the existing target network in the DQN architecture was assumed as a natural candidate in place of the second value function, without having to introduce additional weights or networks. To put it simply, in [1] it is proposed to perform greedy policy through the online (or policy) network and considering the target network to estimate corresponding action value. In reference to both Double Q-learning and DQN, the resulting algorithm is referred to as DDQN. The updating process of DDQN is the same as that of DQN except for replacing the target value function of DQN with (3).

$$Y_t^{DoubleDQN} \equiv R_{t+1} + \gamma Q\left(S_{t+1}, \underset{a}{\operatorname{argmax}}\, Q(S_{t+1}, a; \theta_t); \theta_t'\right) \quad (3)$$

In such settings of DDQN, the target network's update remains identical to that of DQN, while the weights of the target network $\theta_t^-$ for the evaluation of the current greedy policy represent second set of weights introduced in Double Q-learning.

Despite notable improvements offered by DDQN, in the proposed formula of target values in DDQN (3), the parameters $\theta_t$ of online function (i.e., policy network in the case of DDQN) once again emerged which had been withdrawn in the DQN algorithm with the aim of addressing the issue of moving targets in the leaning process of neural networks, which resulted in the introduction of target networks in conjunction with online networks. As a result, here, it is shown how even a slight change in the function of action selection may significantly alter the obtained target value, which in return results in moving targets. Accordingly, in this paper, some possible modifications to DDQN model are presented and discussed.

### *A. Motivation*

The blossom of powerful GPUs in recent years along with the moving targets issue mentioned above, also considering the argument of 'minimal possible changes for handling computational overhead' stated in [1], motivated us into employment of multiple trainable functions (i.e., networks in our case) similar to Double Q-learning and a periodic copy of them at specific steps similar to DQN.

Moreover, stability is of major concern in control in a broad sense, which has largely remained an open question in Reinforcement Learning (RL) [5]; importantly, stability is considered the key to achievement in reinforcement learning in general. Although DQN and DDQN has improved stability through the notion of target network, replay memory, and decoupling of action selection and evaluation (exclusively in DDQN), still any update into policy network could alter the target value due to the fact that action selection is based on policy network, and action value distribution may still be changed with every update. Overall, any achievement in enhancement of stability is of high value.

As an empirical example of mentioned instability occurring in algorithms of DQN and DDQN (although remarkably to a lesser extent in the latter), in many games, including 'Moon Lander' or 'Cart Pole', repeatedly observing the moving average score (over the last 100 episodes) with a continuous upward trend being followed by a downward trend is a common phenomenon; seemingly, the agent suddenly triggers forgetting what it has been learnt up until that moment regardless of chosen hyperparameters or duration of training. Therefore, introducing a more stable target function appears to be a reasonable approach to alleviate this issue, an idea leading to the contributions of this paper.

As a result, here, some modifications to DDQN target value function are introduced. At first, general descriptions of suggested modifications are provided and they will be evaluated more in depth in the next sections.

The first modification, Triple-DQN (TDQN), proposed without any additional computational cost with one trainable network as in DDQN, attempting to attenuate moving target issue and instability in DDQN by offering two target networks, as opposed to one in DDQN, each exclusively dedicated to action selection or action evaluation, the weights of which are periodic copies of policy network at different episode intervals of $N$ and $N/2$.

Two other modifications, namely Semi-decoupled DQN (SD-DQN) and Fully-decoupled DQN (FD-DQN), are proposed each with two and three trainable networks, respectively. SD-DQN can be viewed as a deep network implementation of Double Q-learning algorithm in DQN fashion with two policy networks and corresponding target networks, to assume two sets of target value functions besides decoupling action selection and evaluation. Also, FD-DQN may be viewed as a more complex update to Double Q-learning which employs three policy networks and their corresponding target networks, again, to assume proper number of target value functions and decoupling of action selection and evaluation just similar to SD-DQN, although in one lower level of target value functions' inter-dependence, with the aim of almost fully decouple the target and policy functions compared to SD-DQN which semi-decouples the respective functions, as is the case in Double Q-learning as well.

### *B. Terms' Definition*

To avoid probable ambiguities, the definition of commonly used terms and abbreviations of the paper are provided below: (More in-depth descriptions are discussed in section 3.)

- Primary Target Network: Here, primary target network refers to the target network, which was originally introduced by DQN, however only with the role of 'Q-value evaluation of best action selected' just similar to the target network implemented in target value function of DDQN; the role of 'best action selection' is leaved to another target network, namely 'secondary target network', discussed below.
- Secondary Target Network: Secondary target network, is the target network introduced for 'best action selection' in target value function. In contrast to DDQN, in our newly introduced models, the Q-value evaluation of the selected action is assigned to primary target network, as previously discussed.
- Joint Target Network: The joint implementation of primary and secondary target network is named joint target network. To put it simply, the role of joint target network is to first select the best action through primary target network, then, evaluate the corresponding Q-value utilizing the secondary target network.
- DDQN: DDQN stands for Double-DQN.
- TDQN (Triple Deep Q-network): TDQN stands for Triple-DQN which is named after Double-DQN since overall there are three networks, one trainable policy network plus two untrainable target networks.
- SD-DQN (Semi-decoupled Deep Q-network): SD-DQN stands for Semi-decoupled DQN, since, although not fully decoupled, two trainable separate policy networks are employed as value functions exactly the same as that of Double Q-learning implemented in DQN fashion.
- FD-DQN (Fully-decoupled Deep Q-network): FD-DQN stands for Fully-decoupled DQN, as through employment of three trainable, separate policy networks, a more decoupled model in terms of value functions' interdependency is introduced.

A more detailed and in-depth description along with the corresponding training process of each newly introduced schemas are presented in sections 3 and 4.

The remainder of the paper is structured as follows: section 2 presents a theoretical evaluation of the impact of Mutual Target Network schema on the learning process through drawing parallels with the theoretical assessment of overestimation in the DDQN paper. section 3 outlines the formulations of the newly introduced schemas. section 4 details the results obtained from the conducted experiments, providing insights into their performance. Finally, section 5 concludes the paper with a discussion on the findings and suggestions for future direction.

## II. Mathematical Simulation of Moving Targets

To evaluate the viability of our arguments explained thus far regarding moving target issue, we resort to mathematical simulation carried out in Double-DQN paper [1], which relates to figure 2 of their paper. Similarly, the evaluations are visually illustrated to offer more insight into the proposed arguments. The settings in our experiments are precisely the same as that of theirs with the following details adopted from their paper [1]. To lay the groundwork for our proposed experiments, we first discuss and reimplement the experiments conducted in paper [1], then, we proceed to discuss our supplementary experiments by building it upon their experiments.

In their mathematical simulation, the state space of the problem was considered as a real-valued continuous space where at each state 10 discrete actions are feasible. Also, for the sake of simplicity, it was assumed that the true optimal action values (i.e., $Q_*(s,a)$) depend only on state, that is, at each state, all actions have the same true value which is equal to $V_*(s)$. In each of their experiments, a known function served as the true value function, i.e., either $Q_*(s,a) = sin(s)$ or $Q_*(s,a) = 2\,exp(-s^2)$. Function approximations (representing action value learning) for each of the experiments is done through estimating a d-degree polynomial fit to the true values at sampled states, where d = 6 (top and middle rows) or d = 9 (bottom row). Indeed, they carried out three separate experiments where, in the first experiment they considered 6-D polynomial fitting for $Q_*(s,a) = sin(s)$ function, and for the second and third experiments they considered 6-D and 9-D polynomial fitting for $Q_*(s,a) = 2\,exp(-s^2)$ function, respectively. In their experiment, they considered integer states and their true values (13 integer states in total from -6 to +6) as sampled states whereby, in each of the experiments, to simulate slightly different $Q(s,a)$ per each of the 10 actions, they considered 10 different subsets (through selecting 11 integers from 13 integers mentioned in a way that for each action two adjacent actions were missing while integers -6 and +6 were always included in the subsets to avoid extrapolations) of those sampled states so that each subset corresponds to an action in order to approximate corresponding $Q(s,a)$. In such settings, different functions represent different environments while different fitting polynomials, being 6-D or 9-D, represent different modeling flexibility. In [1] they argue that in many ways this is a typical learning setting, where at each point in time we only have limited data. At last, the outcome of their experiments were illustrated in figure 2 of their paper [1], showcasing how $\max_a Q(S,a;\theta_t)$ operator may be conducive to overestimation regardless of flexibility of the estimator or complexity of the environment.

In order for us to precisely build our experiments with all aforementioned details on that of theirs, we opted to reimplement their experiments, with the results illustrated in Fig. 1. As it can be visually figured out from Fig. 1, the generated figures are exactly the same as the ones in DDQN paper. The same as their figure, each row of the Fig. 1 corresponds to each of the three experiments; the top row relates to 6-D polynomial estimation of $sin(s)$ and middle and bottom rows relates to 6-D and 9-D polynomial estimation of $2\,exp(-s^2)$, respectively. Each of the left column plots show ground truth action value function $Q_*(s,a)$ or $V_*(s)$ (purple line) and a polynomial estimation of action value function $Q(s,a)$ corresponding to a single action (green line) based upon respective sampled states, shown with green dots. Also, the plots in middle column shows polynomial estimation of action value functions for all 10 actions (green lines) as functions of state, along with the maximum action value at each state $\max_a Q(S,a)$ (black dashed curve). It should be noted that each action's function approximation (each of the green lines), i.e., $Q(s,a)$, differs while the true value function is the same for all actions since different sets of sampled states were considered for each action. The right column's plots indicate the difference between the black dashed curve and purple curves in orange (i.e., $\max_a Q(S,a) - V_*(s)$). The right plots also show the Double Q-learning estimate error in blue (i.e., $Q'(S, argmax_a Q(s,a)) - V_*(s)$ ). To calculate Double Q-learning estimate error of right column's plots, the double estimators of each action value function corresponding to action $a_i$, the action value function estimation corresponding to either action $a_{i+5}$ or $a_{i-5}$, were considered if $i \leq 5$ or $i > 5$, respectively.

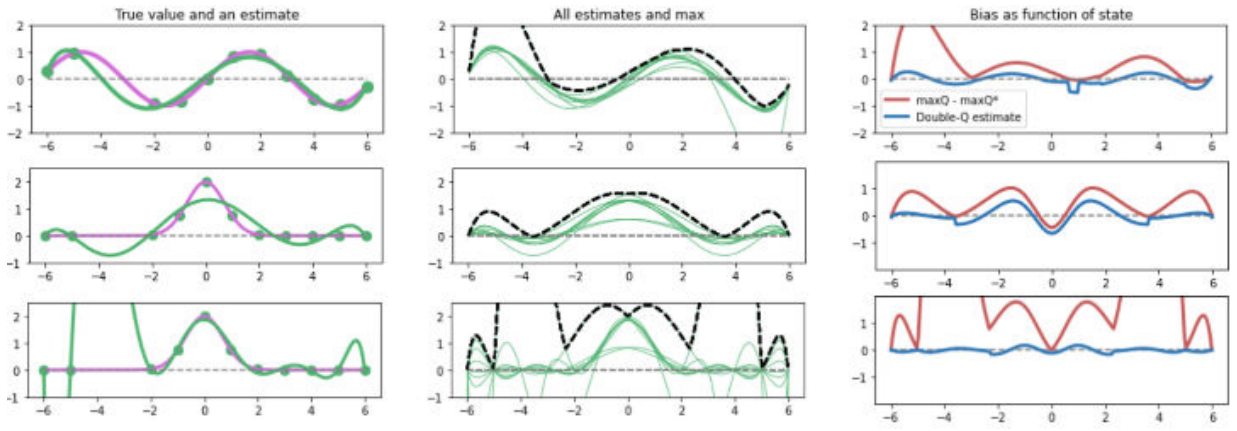

Fig. 1. Reimplementation of Function Approximation Experiment of [1]

In their paper [1], they argue that such selection for double estimator had been arbitrarily; however, as a side note, here we argue that, in such setting, within valid range for $i$, that is 0 to 5, any number other than 5 results in poorer results; indeed, as we lower $i$ from 5 to 0, the more Double Q-learning estimate (here, denoted as $DQE_i$) reflects black dashed curve, that is $\max_a Q(S,a)$. Here, we have illustrated the Double Q-learning curves for other values of $i$ (here, denoted as $j$) for 0 to 4 for all three experiments in Fig. 2Fig. 3,Fig. 4. Also, the sum of squared errors (SSE) for Double Q-learning (i.e., $Q'(S, argmax_a Q(s,a))$ against $V_*(s)$) for $i = 5$ estimates are given in TABLE I.

Indeed, through our experiments, we are to simulate how policy network updates may result in fluctuations in action value estimations regarding Double Q-learning algorithm. Also, the implications of estimator flexibility and environment complexity will be analyzed as well.

To do so, we have simulated an update to policy network with substitution of the double estimator of index $i$ (denoted as $Q_i(s,a)$) with double estimator of index $j$ (denoted as $Q_j(s,a)$), to put it simply, by changing the value of $i$. Therefore, Fig. 2Fig. 3,Fig. 4 illustrate Double Q-learning estimates for $j = 0,1,2,3,4$ ($DQE_j$) on top of Double-Q learning estimate for $i = 5$ ($DQE_i$). Note that values of '$i$' other than 5 are denoted as '$j$' in our figures.

TABLE I. SUM OF SQUARED ERRORS

| Function | SSE |
| --- | --- |
| $Sin(x)$ w. 6-D | 6.55 |
| $2e^{(-x^2)}$ w. 6-D | 16.30 |
| $2e^{(-x^2)}$ w. 9-D | 1.34 |

As it can be figured out from each of the figures below, as indexes of '$i$' and '$j$' become further away, the corresponding SSE (specified above each plot) of $DQE_i$ against $DQE_j$ almost becomes greater. More importantly, except for second experiment (Fig. 3), in other two experiments, SSEs of $DQE_i$ against $DQE_j$ are multiple times greater than SSEs of $DQE_i$ against $V_*(s)$ given in TABLE I. Note that the only difference here lies in the value of $i$ in $DQE_i$ (i.e., best action selection function for Double Q-learning).

What is critical here is that with even slight changes in the distribution of action selection function the obtained errors may drastically change, thereby contributing to moving targets issue, which may deteriorate the learning process.

Considering Fig. 3, except for the first plot (i.e., $i = 5$ and $j = 0$ ), SSEs do not differ markedly compared to the corresponding SSE of 16.3 given in TABLE I. This is sensible and could be expected considering the distribution of action value function estimations $Q(s, a)$, as is illustrated second plot of second row in Fig. 1, showcasing action value functions' distribution contributing factor in moving targets issue.

Moreover, considering Fig. 4, which is the most similar experiment compared to the other two to real-world scenarios since in this experiment flexibility of neural networks reflects in more flexible 9-D estimator as opposed to 6-D. As it can be noticed in Fig. 4, slight alterations in action selection estimators result in extremely higher squared errors. Hence, by comparison, this highlights that the issue of moving targets may be largely dependent on flexibility of models in hand as well.

In summary, through such simulation it is underscored how even slight changes or any updates to policy in general may result in dramatic fluctuations in target values approximations, which in return underlies moving targets issue in real world scenarios; therefore, addressing this issue seems of high value.

## III. FORMULATION OF DIFFERENT SCHEMAS

In the following, newly introduced schemas' formulations along with their corresponding descriptions and learning process are discussed.

### A. TDQN

TDQN is an update to DDQN with only one modification, that is deployment of secondary target network, employed together with the existed primal target network. The secondary target network is a copy of policy network after each $N/2$ episodes (or steps in general), assuming $N$ as the number of episodes after which the weights of policy network are copied to primal target network, thereby we expect to further address the moving targets issue and reach to a more stable model and training. The process of training the TDQN is exactly the same as DDQN except that the best action is to be selected through the secondary target network, i.e., $Q_{\theta''}$

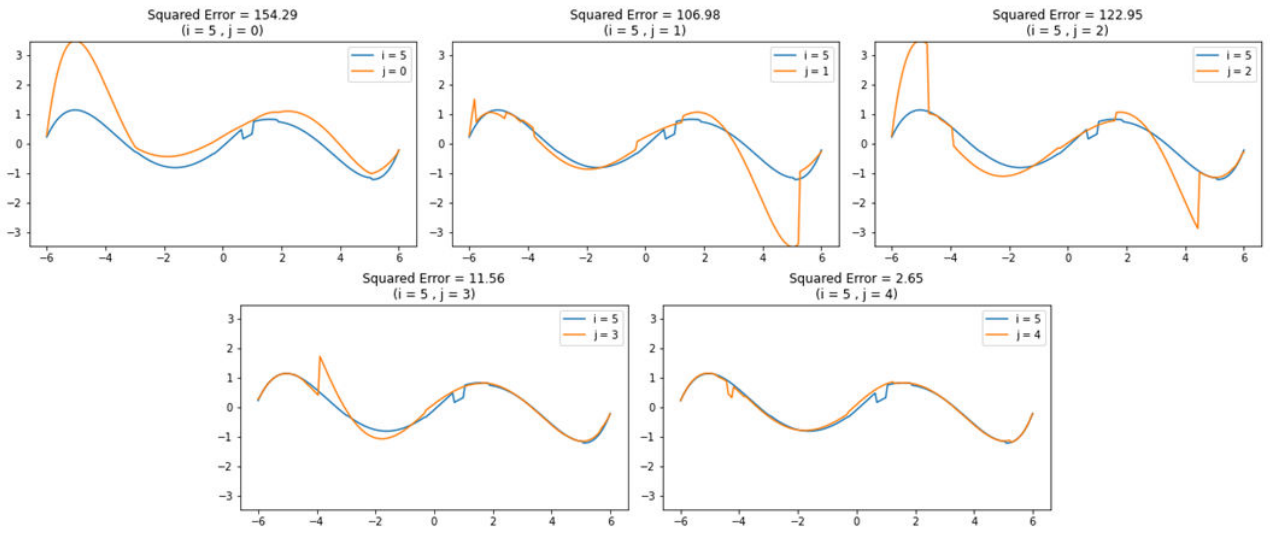

Fig. 2. Double Q-learning Estimates $DQE_{i\ or\ j}$ for the Function $Sin(x)$ with 6-D Polynomial Estimation

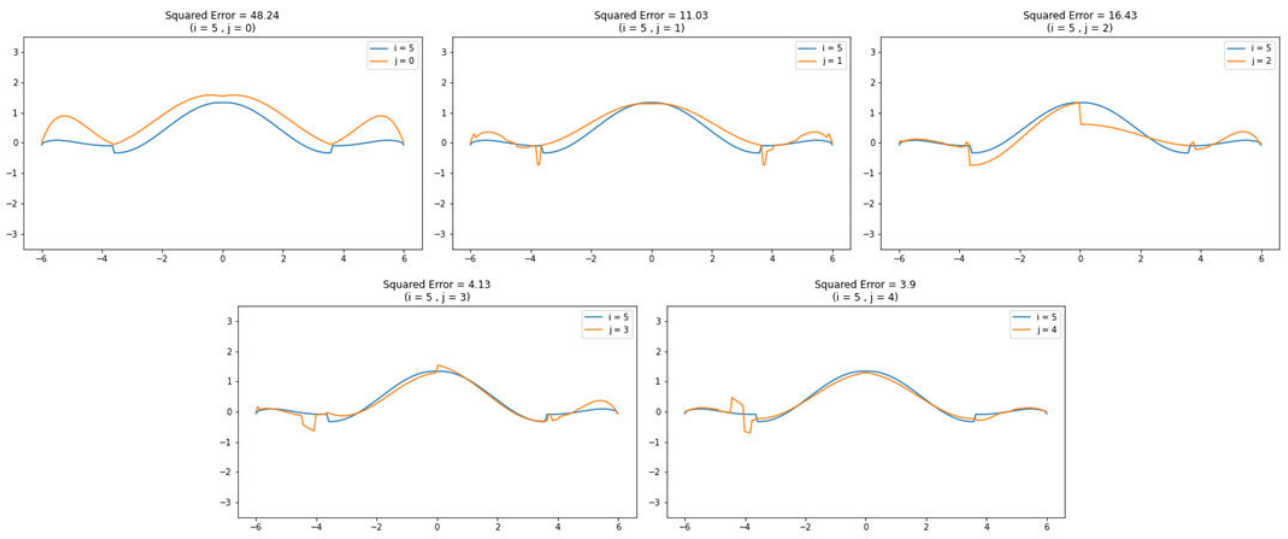

Fig. 3. Double Q- learning Estimates $DQE_{i\ or\ j}$ for the Function $2exp(-x^2)$ with 6-D Polynomial Estimation

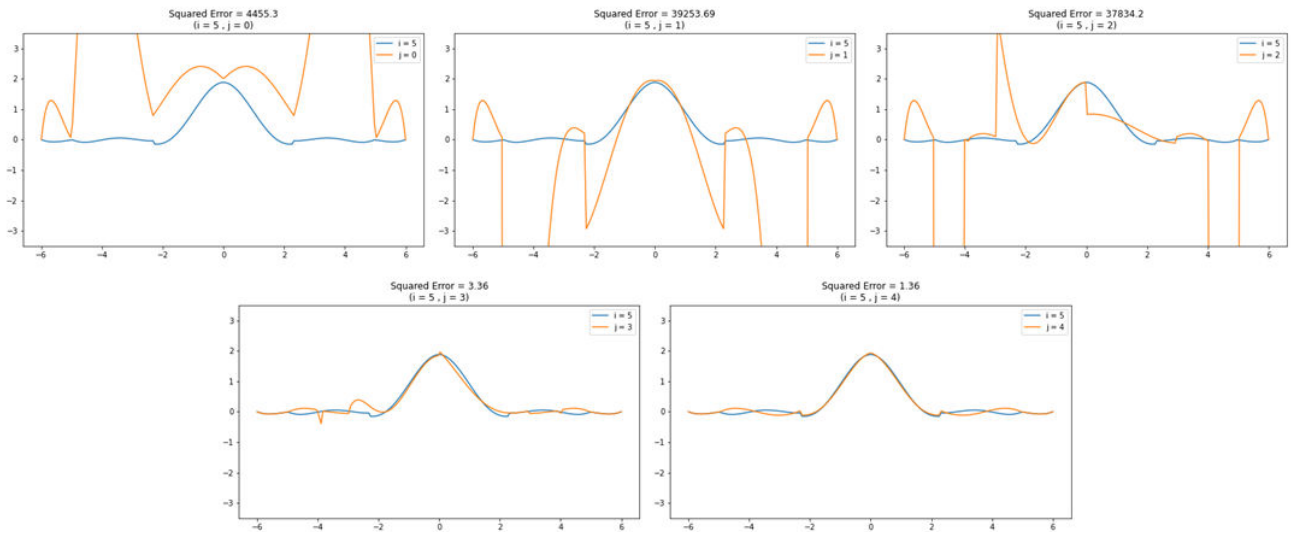

Fig. 4. Double Q- learning Estimates $DQE_{i\ or\ j}$ for the Function $2exp(-x^2)$ with 9-D Polynomial Estimation

with regard to the corresponding target value function of TDQN (4).

$$Y_t^{TDQN} \equiv R_{t+1} + \gamma Q\left(S_{t+1}, \underset{a}{argmax}\ Q(S_{t+1}, a; \theta_t''), \theta_t'\right) \quad (4)$$

### B. SD-DQN

Further adapting DDQN towards Double Q-learning model in DQN fashion, in resulting model SD-DQN two networks (i.e., two sets of weights) are interdependently trained as Q-value approximators, where each of these networks would have its corresponding primary target network. Hence, overall, there are two policy networks along with two corresponding primary target networks. The logic behind this formulation is to practically combine the benefits of all mentioned settings so far; that is, SD-DQN not only benefits from double value functions to overcome overestimations, but it also benefits from target network introduced by DQN besides decoupling logic of DDQN, and moreover, it also takes advantage of TDQN by employment of primary target network. Therefore, by this setting we expect further stabilization of the model.

The process of learning is exactly similar to Double Q-learning since in SD-DQN both functions are updated in parallel by randomly assigning each experience from replay buffer to each of the functions.

$$\begin{cases} Y_t^1 \equiv R_{t+1} + \gamma Q_2\left(S_{t+1}, \underset{a}{\operatorname{argmax}} Q_1\left(S_{t+1}, a; \boldsymbol{\theta}'_{t_1}\right), \boldsymbol{\theta}'_{t_2}\right) \\ Y_t^2 \equiv R_{t+1} + \gamma Q_1\left(S_{t+1}, \underset{a}{\operatorname{argmax}} Q_2\left(S_{t+1}, a; \boldsymbol{\theta}'_{t_2}\right), \boldsymbol{\theta}'_{t_1}\right) \end{cases} \quad (5)$$

### C. FD-DQN

FD-DQN is actually a revision to SD-DQN where it is further decoupled. Despite its naming as Fully Decoupled-DQN, it has actually not been fully decoupled which may be inferred by untangling value functions through substituting respective nested Q-value functions.

$$\begin{cases} Y_t^1 \equiv R_{t+1} + \gamma Q_2\left(S_{t+1}, \underset{a}{\operatorname{argmax}} Q_3\left(S_{t+1}, a; \boldsymbol{\theta}'_{t_3}\right), \boldsymbol{\theta}'_{t_2}\right) \\ Y_t^2 \equiv R_{t+1} + \gamma Q_3\left(S_{t+1}, \underset{a}{\operatorname{argmax}} Q_1\left(S_{t+1}, a; \boldsymbol{\theta}'_{t_1}\right), \boldsymbol{\theta}'_{t_3}\right) \\ Y_t^3 \equiv R_{t+1} + \gamma Q_1\left(S_{t+1}, \underset{a}{\operatorname{argmax}} Q_2\left(S_{t+1}, a; \boldsymbol{\theta}'_{t_2}\right), \boldsymbol{\theta}'_{t_1}\right) \end{cases} \quad (6)$$

The process of learning in FD-DQN is exactly the same as that of SD-DQN except for three value function estimators.

## IV. PERFORMANCE COMPARISON

In this section, we evaluate the performance of each introduced schema in practice. To do so, two different environments of OpenAI Gym [6] were selected, 'CartPole_v0' game from the 'classic control environment' and 'Pong_v4' game from the 'Atari environment'.

The goal of these experiments was not to solve games, fine-tune hyperparameters, or custom tailor network's architectures of the introduced schemas but to fairly evaluate and compare the performance of each proposed schema and exploring probable enhancements in stability especially for a sensible and affordable number of episodes. The performance of DDQN had served as our benchmark in the experiments. It should be noted that chosen hyperparameters were already fine-tuned for DDQN; in this case, introduced schemas were not necessarily expected to outperform DDQN, but our main objective was to explore whether stability of trainings were improved in the process of learning. The utilized hyperparameter were obtained from [7] for 'Cartpole' and [8] for 'Pong'.

For 'Cartpole' two architectural settings were considered. First setting was implemented with neural network with three linear layers followed by 'ReLU' activation functions [9]. And the second was implemented with deeper neural network with three convolutional layers with 'ReLU' activation functions followed by two linear layers. The second setting is actually the setting previously offered by DDQN paper [1]. Lastly, for the game 'Pong' only one setting which is exactly similar to the second setting of 'Cartpole' was considered. Before diving into the obtained outcomes, it should be noted that these experiments were conducted under extremely limited hardware resources.

In the following, obtained outcomes for each part are illustrated. It should be noted that illustrated outcomes are obtained after multiple runs for the sake of credibility. All the following figures include outcomes of all schemes' outputs. It should be noted that each type of game has its respective output (y-axis) depending on its type; for instance, the outputs in the game 'Cartpole' denotes the duration of each episode before failing (the greater the better) while the outputs in the game 'Pong' denotes the obtained score at the end of each episode (the greater the better as well).

### A. 'CartPole' with Linear Layers

In this experiment, the models were implemented with linear layers for the game 'Cartpole', while each model was trained for up to 1500 episodes. As shown in Fig. 5 for one sample of runs, in terms of both 'episode duration' and 'stability', the model FD-DQN has outperformed all other models in almost all runs at both earliest and latest stages; also, it worths mentioning that the model TDQN has shown the smoothest trend during training specifically throughout first 1000 episodes. Each model's corresponding execution time is also shown in Fig. 5.

### B. 'CartPole' with Convolutional Layers

In this experiment, the models were implemented with convolutional layers for the game 'Cartpole'. This setting was executed for 1500 and 2000 episodes, with their results shown in Fig. 6 and Fig. 7, respectively. In almost all runs of this setting, both the models SD-DQN and FD-DQN have expressed a better performance than the DDQN model in both optimal policy learning and training stability, while the model TDQN had an on-par performance compared to the DDQN model. Each model's corresponding execution time is also shown in Fig. 6 and Fig. 7.

### C. 'Pong' with Convolutional Layers

In this experiment, the models were implemented with convolutional layers for the game 'Pong', while each model was trained for up to 500 episodes. In this settings, surprisingly, both 'SD-DQN' and 'FD-DQN' appeared to collapse and act randomly without any upward trend or signs of learning; however, the underlying factor to this phenomenon requires to be examined in more depth in order to ensure correctness of implementations for this specific game; regardless, in this setting, the model TDQN had remarkably outperformed DDQN in both stability and learning optimal policy for multiple runs as it can be figured out from Fig. 8, TDQN has remarkably more stable upward trend as against DDQN.

In summary, it can be inferred that in almost all of the conducted experiments, at least one of the introduced schemas have illustrated more stability and optimal policy learning capacity compared to DDQN model in the horizon of respective number of episodes; that is, the model TDQN has illustrated remarkably more stability and policy learning capacity in the game 'Pong' while models SD_DQN and FD-DQN have expressed more stability and policy learning capacity in the game 'Cartpole'. More importantly, all these outcomes were obtained with hyperparameters having been fine-tuned for model DDQN; additionally, presented findings are consistent with the theoretical explanations discussed in section 2.

## V. CONCLUSION

In this paper, we have presented three main contributions to modify the model Double-DQN. First, we discussed and illustrated mathematically how action selection based on policy network may result in moving targets issue, and how different environments and settings may critically exacerbate the problem. Second, we have introduced three schemas with modifications to the existing algorithms regarding decoupling of action selection and evaluation to further tackle the instability of learning process existing in both Double-DQN and DQN models. Third, we have shown that even with hyperparameters having been fine-tuned for Double-DQN model, in almost all of the conducted experiments, at least one the newly introduced models outperformed DDQN in terms of stability and optimal policy learning. Finally, in order to further research in the scope of this paper, one may evaluate the efficacy of averaging, minimizing or maximizing over all learned policy networks (particularly in schemas with more than one policy network) in order to further overcome possible overestimations or underestimations that may have been occurred in any of the introduced schemas; this provides more flexibility when encountering different environments.

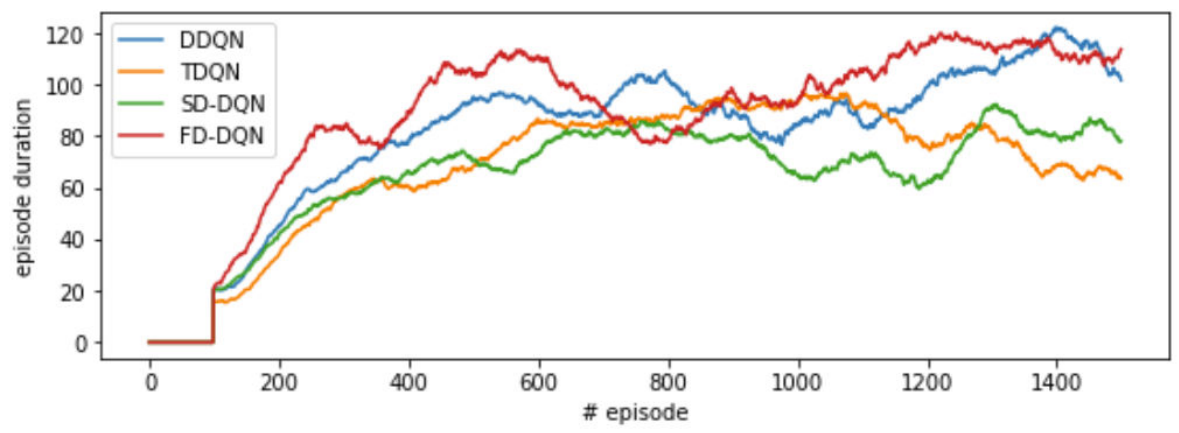

Fig. 5. Cartpole with Linear Layers Trained for 1500 Episode

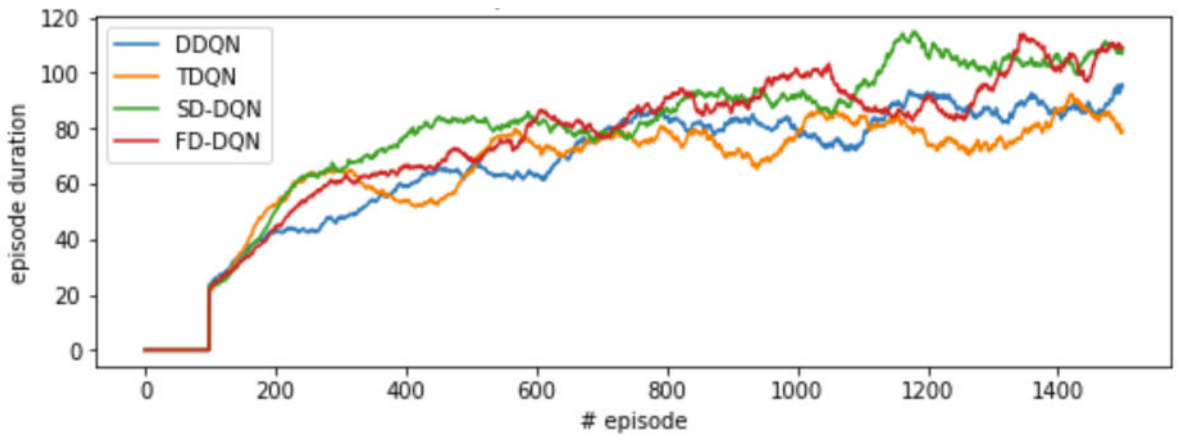

Fig. 6. Cartpole with Convolutional Layers Trained for 1500 Episodes

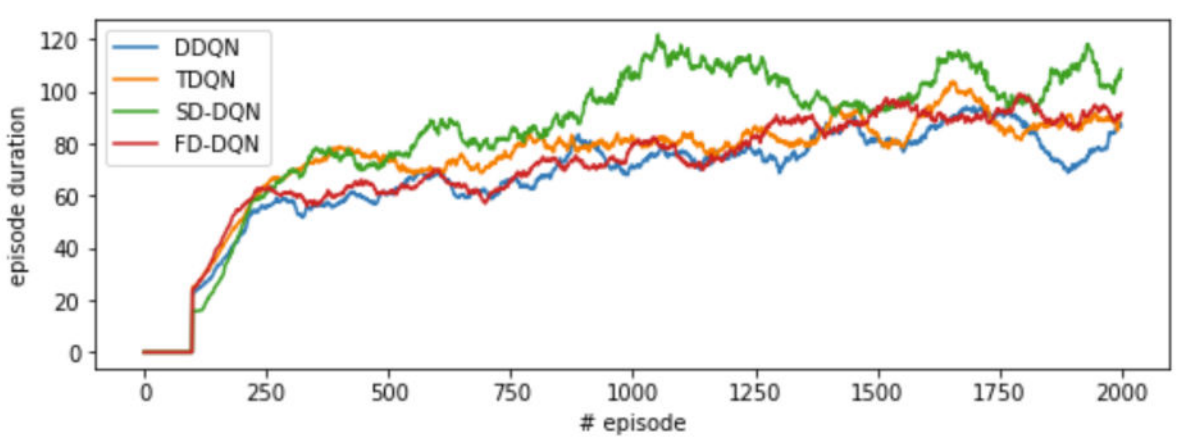

Fig. 7. Cartpole with Convolutional Layers Trained for 2000 Episodes

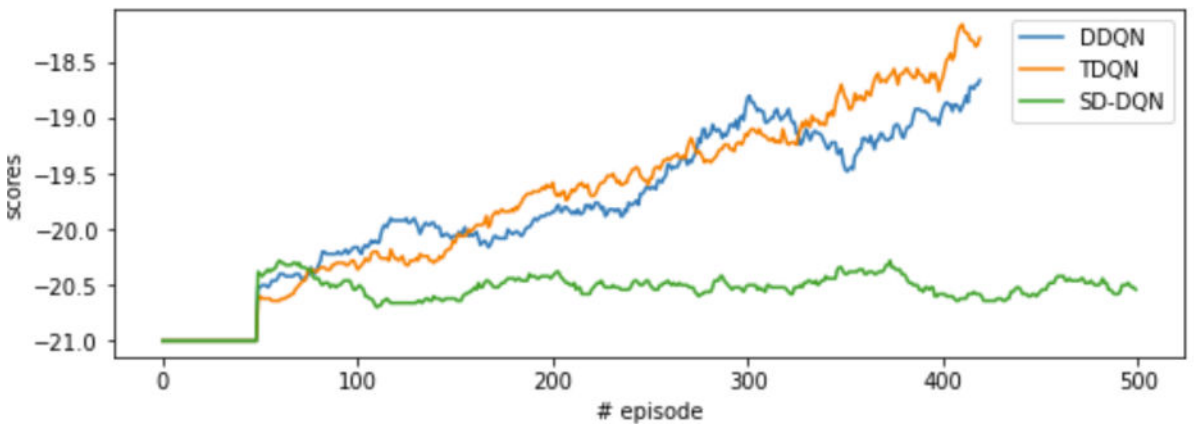

Fig. 8. Pong with Convolutional Layers Trained for 500 Episodes

## REFERENCES

[1] H. Van Hasselt, A. Guez, and D. Silver, "Deep reinforcement learning with double q-learning," in Proceedings of the AAAI conference on artificial intelligence, 2016, vol. 30, no. 1.

[2] H. Hasselt, "Double Q-learning," *Advances in neural information processing systems,* vol. 23, 2010.

[3] V. Mnih, "Playing atari with deep reinforcement learning," *arXiv preprint arXiv:1312.5602,* 2013.

[4] V. Mnih *et al.*, "Human-level control through deep reinforcement learning," *nature,* vol. 518, no. 7540, pp. 529-533, 2015.

[5] L. Buşoniu, T. De Bruin, D. Tolić, J. Kober, and I. Palunko, "Reinforcement learning for control: Performance, stability, and deep approximators," *Annual Reviews in Control,* vol. 46, pp. 8-28, 2018.

[6] G. Brockman *et al.*, "Openai gym," *arXiv preprint arXiv:1606.01540,* 2016.

[7] A. Green. "DDQN Hyperparameter Tuning Using Open AI Gym Cartpole." ADG Efficiency. https://adgefficiency.com/dqn-tuning/.

[8] J. Michaux. "DQN PyTorch." Github. https://github.com/jmichaux/dqn-pytorch.

[9] A. Agarap, "Deep Learning Using Rectified Linear Units (ReLU)," *arXiv preprint arXiv:1803.08375,* 2018.